\documentclass[conference]{IEEEtran} 
\IEEEoverridecommandlockouts
\usepackage{cite}
\usepackage{amsmath,amssymb,amsfonts}
\usepackage{algorithmic}
\usepackage{graphicx}
\usepackage{placeins}
\usepackage{subcaption}
\usepackage{textcomp}
\usepackage{xcolor}
\usepackage{hyperref}       
\usepackage{url}            
\usepackage{booktabs}       
\usepackage{amsfonts}       
\usepackage{nicefrac}       
\usepackage{microtype}      
\usepackage{xcolor}         
\usepackage{tikz}
\usetikzlibrary{shapes.geometric,positioning, backgrounds, calc, arrows,fit}

\newcommand{\alg}{{$\textsc{LAVAE}$}}

\def\BibTeX{{\rm B\kern-.05em{\sc i\kern-.025em b}\kern-.08em
    T\kern-.1667em\lower.7ex\hbox{E}\kern-.125emX}}

\tikzset{arrow/.style={-stealth, thick, draw=gray!80!black}}

\newcommand\denselyConnectNodes[2]{
  \foreach \n [count=\lyrIdx, remember=\lyrIdx as \previdx, remember=\n as \prevn] in #2 {
    \foreach \y in {1,...,\n} {
      \ifnum \lyrIdx > 1
        \foreach \x in {1,...,\prevn}
          \draw[->] (#1-\previdx-\x) -- (#1-\lyrIdx-\y);
      \fi
    }
  }
}

\begin{document}

\title{Towards Composable Distributions of Latent Space Augmentations\\
}

\author{
\IEEEauthorblockN{Omead Pooladzandi$^{\dagger}$}
\IEEEauthorblockA{
opooladz@ucla.edu}
\and
\IEEEauthorblockN{Jeffrey Jiang$^{\dagger}$}
\IEEEauthorblockA{
jimmery@ucla.edu}
\and
\IEEEauthorblockN{Sunay Bhat$^{\dagger}$
\thanks{$^{\dagger}$ \text{Equal Contribution}}
\thanks{\textit{Department of Electrical and Computer Engineering}}
\thanks{\textit{University of California, Los Angeles}}
}
\IEEEauthorblockA{
sunaybhat1@ucla.edu}
\and
\IEEEauthorblockN{Gregory Pottie$^{\dagger}$}
\IEEEauthorblockA{
pottie@ee.ucla.edu}
}

\maketitle

\begin{abstract}

We propose a composable framework for latent space image augmentation that allows for easy combination of multiple augmentations. Image augmentation has been shown to be an effective technique for improving the performance of a wide variety of image classification and generation tasks. Our framework is based on the Variational Autoencoder architecture and uses a novel approach for augmentation via linear transformation within the latent space itself. We explore losses and augmentation latent geometry to enforce the transformations to be composable and involuntary, thus allowing the transformations to be readily combined or inverted. Finally, we show these properties are better performing with certain pairs of augmentations, but we can transfer the latent space to other sets of augmentations to modify performance, effectively constraining the VAE's bottleneck to preserve the variance of specific augmentations and features of the image which we care about. We demonstrate the effectiveness of our approach with initial results on the MNIST dataset against both a standard VAE and a Conditional VAE. This latent augmentation method allows for much greater control and geometric interpretability of the latent space, making it a valuable tool for researchers and practitioners in the field.

\end{abstract}


\section{Introduction}

Data augmentation has become an essential technique in deep learning, allowing models to learn from a diverse set of input images by applying various types of transformations, such as rotation, flipping, cropping, and color shifting. By artificially increasing the size of the training dataset, data augmentation can reduce the risk of overfitting and improve the generalization of the model to new, unseen data.

However, choosing the right set of augmentation techniques for a given task can be challenging. Some types of augmentations may affect the way an image is interpreted, such as flipping or rotating digits in handwritten digit recognition tasks. Practitioners must employ priors on the data to know which augmentations are appropriate.

In this paper, we introduce a novel approach for latent image augmentation using Variational Autoencoder (VAE) architecture. Our approach allows for the easy combination of multiple augmentation techniques and provides greater control and interpretability of the latent space. Within the latent space of the VAE architecture, we can apply, compose, and invert linear transformations to generate augmented versions of the input images. The key contribution of our work is the use of latent-space linear transforms and a two-step training method to learn mappings between the original and augmented latent spaces, with a surprising emergence of composability. We further demonstrate that our approach can transfer a trained latent space to a new set of augmentations using a multiple decoder architecture, enabling practitioners to transfer certain properties and potential performance improvements dependent on the original augmentations.

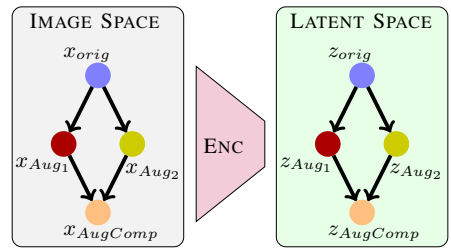
\begin{figure}[h]
    \centering
    \resizebox{0.37\textwidth}{!}{\begin{tikzpicture}

    \draw[rounded corners, fill=gray!10] (-0.75, -2.5) rectangle (1.75, 1) {};
    
    \node[draw, shape=circle, color=blue!50, fill=blue!50, very thick, minimum size = 0.02cm] at (0.5,0) (x_og) {};
    \node[draw, shape=circle, color={rgb:black,1;red,2}, fill={rgb:black,1;red,2}, very thick, minimum size = 0.02cm] at (0, -1) (x_aug1) {};
    \node[draw, shape=circle, color={rgb:black,1;yellow,4}, fill={rgb:black,1;yellow,4}, very thick, minimum size = 0.02cm] at (1, -1) (x_aug2) {};
    \node[draw, shape=circle, color=orange!50, fill=orange!50, very thick, minimum size = 0.02cm] at (0.5, -2) (x_comp) {};

    \node [text width=3cm] (0) at (1, 0.8) {\textsc{\small{Image Space}}};
    \node [text width=1cm] (0) at (0.5, 0.3) {\small $x_{orig}$};
    \node [text width=1cm] (0) at (-0.2,-1.3) {\small $x_{Aug_1}$};
    \node [text width=1cm] (0) at (1.4,-1.4) {\small $x_{Aug_2}$};
    \node [text width=1cm] (0) at (0.5,-2.3) {\small $x_{Aug Comp}$};

    \draw[ultra thick, ->] (x_og) -- (x_aug1);
    \draw[ultra thick, ->] (x_og) -- (x_aug2);
    \draw[ultra thick, ->] (x_aug1) -- (x_comp);
    \draw[ultra thick, ->] (x_aug2) -- (x_comp);

    \draw[fill=purple!20] ([xshift=1.3cm]x_og.north east) -- ([xshift=2.3cm,yshift=-0.8cm]x_og.north east) -- ([xshift=2.3cm,yshift=0.8cm]x_comp.south east) -- ([xshift=1.3cm]x_comp.south east) -- cycle; 
    \node (ENC) at ([xshift=1.7cm,yshift=-0.9cm]x_og.south east) {\textsc{\small{Enc}}};

    \begin{scope}[shift={(3.85,0)}]
        \draw[rounded corners, fill=green!10] (-0.75, -2.5) rectangle (1.75, 1) {};
    
        \node[draw, shape=circle, color=blue!50, fill=blue!50, very thick, minimum size = 0.02cm] at (0.5,0) (z_og) {};
        \node[draw, shape=circle, color={rgb:black,1;red,2}, fill={rgb:black,1;red,2}, very thick, minimum size = 0.02cm] at (0, -1) (z_aug1) {};
        \node[draw, shape=circle, color={rgb:black,1;yellow,4}, fill={rgb:black,1;yellow,4}, very thick, minimum size = 0.02cm] at (1, -1) (z_aug2) {};
        \node[draw, shape=circle, color=orange!50, fill=orange!50, very thick, minimum size = 0.01cm] at (0.5, -2) (z_comp) {};

        \node [text width=3cm] (0) at (0.95, 0.8) {\textsc{\small{Latent Space}}};
        \node [text width=1cm] (0) at (0.5, 0.3) {\small $z_{orig}$};
        \node [text width=1cm] (0) at (-0.2,-1.3) {\small $z_{Aug_1}$};
        \node [text width=1cm] (0) at (1.4,-1.4) {\small $z_{Aug_2}$};
        \node [text width=1cm] (0) at (0.5,-2.3) {\small $z_{Aug Comp}$};
    \end{scope}

    \draw[ultra thick, ->] (z_og) -- (z_aug1);
    \draw[ultra thick, ->] (z_og) -- (z_aug2);
    \draw[ultra thick, ->] (z_aug1) -- (z_comp);
    \draw[ultra thick, ->] (z_aug2) -- (z_comp);
    
\end{tikzpicture}}
    \caption{Approximate model/DAG \textit{learned} in latent space for \textit{known} image space augmentations}
    \label{fig: DAG}
\end{figure}

Our experiments on the MNIST dataset demonstrate that our proposed approach can improve the performance of VAEs and provide new insights into the underlying structure of the data and the relationship between different augmentations. By viewing augmentations as image-space priors and not data to simply be randomized across, we can constrain the VAE's information bottleneck and improve its generalization ability. In essence, our method learns a low-dimensional, latent-proxy \ref{fig: DAG} for a set of image-space functions, even when the image space model or transformation process is unknown a priori, as long as training samples exist of the augmented images.
\section{Related Work}

Control and manipulation of a lower-dimensional latent space in generative modeling is an area of ongoing research. The Conditional VAE (CVAE) is an initial extension of a vanilla VAE in which a conditional value or ``one-hot'' encoding is concatenated to both the Encoder and Decoder inputs \cite{sohn_cvae:2015}. The CVAE does a form of latent space separation by adding dimensions based on a conditional variable, and it presents the most compelling comparison to our own work as it allows unique interpretation of the latent space based on a prior or semantic label. Other extensions of VAEs include the VQ-VAE, which uses a discrete latent space to model discrete data types such as text, and the Flow-based VAE, which uses normalizing flows to model complex posterior distributions \cite{aaron_vqvae:2017,Su_fvae:2018}. Latent diffusion models are another approach that iteratively add noise  in the training process and can reverse this process in inference to achieve state of the art text to image and image completion and synthesis tasks \cite{rombach_diff:2022}.

The latent space can also be used to apply lower-dimensional modeling or priors. Causal generative models have seen a variety of success with both learning and utilizing causal information and structural models to generate counterfactual images and datasets \cite{yang_causalvae:2021,kocaoglu_causalgan:2017,bhat_debias:2022,bhat_hypothesis:2022,breugel_decaf:2021}. Our method looks to also extend interpretability of the latent space by approximating image augmentations, a priors-based prep-processing approach in the image-space, in the latent space. This could loosely be thought of as a causal model proxy to the image space causal model, Figure \ref{fig: DAG}.
\section{Background}


\subsection{Variational Autoencoders}

The VAE framework is based on the principle of maximum likelihood estimation, where the goal is to maximize the likelihood of the training data under the model. However, in order to make the optimization tractable, the VAE introduces a variational lower bound on the log likelihood, which can be written as:

\begin{align}
\mathcal{L}(\theta,\phi;x^{(i)}) &= \mathbb{E}_{e_{\phi}(z|x^{(i)})}\left[\log d_\theta(x^{(i)}|z)\right] \\ &-\text{KL}\left(e_\phi(z|x^{(i)})||m_\theta(z)\right) \nonumber \\
&= \mathbb{E}_{e_{\phi}(z|x^{(i)})}\left[\log d_\theta(x^{(i)}|z)\right] \\
&- \int e_\phi(z|x^{(i)}) \log \frac{e_\phi(z|x^{(i)})}{m_\theta(z)} dz\nonumber
\end{align}

where $x^{(i)}$ is a single training example, and $\theta$ and $\phi$ are the parameters of the decoder and encoder, respectively. The first term in the lower bound, $\mathbb{E}_{e_{\phi}(z|x^{(i)})}\left[\log d_\theta(x^{(i)}|z)\right]$, is known as the reconstruction loss, and it measures the difference between the reconstructed data and the original data. The second term, $\text{KL}\left(e_\phi(z|x^{(i)})||m_\theta(z)\right)$, is known as the KL divergence, and it measures the difference between the approximate posterior distribution and the latent distribution. The first term is the decoding error (the classic rate-distortion theory), and the second term is the extra rate for coding $z$ assuming marginal pdf $m_\theta(z)$.

The Conditional VAE (CVAE) is a natural extension of the VAE framework that adds a conditional input to both the encoder and decoder networks. In the CVAE, the goal is to learn a conditional generative model that can generate new samples from a specific class or condition, given some additional information (additional details in Appendix \ref{app: CVAE}). By adding the conditional input to the VAE framework, the CVAE can generate samples conditioned on a specific input, which is useful in many applications. For example, in image generation, given a class label as the conditional input, the CVAE can generate images of that class. 

\subsection{Priors in Pre-processing}

Data augmentation is not done naively, or without a strong sense of priors. In image datasets, typical augmentations might include crops, rotations, flips, scaling, color modifications, masks, and many more. In order to expand the training domain and learn a more robust model, only augmentations which are invariant to the classification or interpretation of the resulting image can be applied, and similarly, negative augmentations which impact classification can be used to refine the support of a distribution \cite{Sinha_negAug:2021}. As a simple example, a left-right flip is an acceptable transformation for a 0 or 8 digit, but not for a 2 or 9. One can think of augmentation as a causal model or directed-graph in the image space, in which all augmented image distributions are the result of applying a transform or function to an parent node of original images such that the resulting images are still within the same class. Succinctly, given the original dataset $D_{orig}$, 
\begin{gather}
    D_{aug} = \bigcup_{\substack{i \in \mathcal{A}_c \\ d \in D_{orig} \\ c = class(d)}} \ f_{aug_i}(d)  \\
    s.t. \quad \mathcal{A}_c = \{ i : class(d) = class( f_{aug_i}(d) );  d \in D_{orig} \} 
    \label{eq:aug_prior}
\end{gather}
for some pre-defined set of augmentations $\{f_{aug_i}\}$. 

This model is typically well known and easily applied in pre-processing, but is not made explicit. Although these augmented distributions are classified or interpreted similarly at a high-level, there are functional relationships and structure between them that we look to make explicit.

\section{Latent Augmentation VAE}

\subsection{Architectural Overview}
\begin{figure*}[!htb]
    \centering
    \resizebox{0.7\textwidth}{!}{

\begin{tikzpicture}

	\node[fill=blue!20, minimum width=1.2cm, minimum height=1.2cm] (X_og) at (0,1) {$\mathbf x_{OG}$};
        \node[fill=blue!20, minimum width=1.2cm, minimum height=1.2cm] (X_aug) at (0,-0.4) {$\mathbf x_{Aug}$};

	\draw[fill=purple!20] ([xshift=0.5cm]X_og.north east) -- ([xshift=2cm,yshift=-0.8cm]X_og.north east) -- ([xshift=2cm,yshift=0.8cm]X_aug.south east) -- ([xshift=0.5cm]X_aug.south east) -- cycle; 
	\node (ENC) at ([xshift=1.2cm,yshift=-0.1cm]X_og.south east) {$e_{\phi}(z|x)$};

	\node[fill=red!50, minimum width=0.5cm, minimum height=1.0cm] (Z) at (3.7cm,-0.2cm) {\tiny $\mathbf z_{OG,Aug}$};
        \node[fill=red!50, minimum width=0.5cm, minimum height=1.0cm] (Z_h) at (6.3cm,1.2) {\tiny $\mathbf {\hat{z}}_{OG,Aug}$};

        \draw[fill=purple!20] (4.1,1.5) -- (4.7,1.9) -- (5.3, 1.5) -- 
                      (5.3,1.1) -- (4.7,0.7) -- (4.1,1.1) -- cycle;
        
        \draw[fill=purple!20] (4.05,1.45) -- (4.65,1.85) -- (5.25, 1.45) -- 
                      (5.25,1.05) -- (4.65,0.65) -- (4.05,1.05) -- cycle;
                      
        \node (Laug) at (4.6,1.2) {\footnotesize \textsc{$L_{aug_{i}}$}};

	\node[fill=blue!20, minimum width=1.2cm, minimum height=1.2cm] (X_hog) at (10,1) {$\mathbf{\hat{x}}_{OG}$};
        \node[fill=blue!20, minimum width=1.2cm, minimum height=1.2cm] (X_haug) at (10,-0.4) {$\mathbf{\hat{x}}_{Aug}$};

	\draw[fill=purple!20] ([xshift=-2cm,yshift=-0.8cm]X_hog.north west) -- ([xshift=-0.5cm]X_hog.north west) -- ([xshift=-0.5cm]X_haug.south west) -- ([xshift=-2cm,yshift=0.8cm]X_haug.south west) -- cycle;
         \draw[fill=purple!20] ([xshift=-2.05cm,yshift=-0.85cm]X_hog.north west) -- ([xshift=-0.55cm,yshift=-0.05cm]X_hog.north west) -- ([xshift=-0.55cm,yshift=-0.05cm]X_haug.south west) -- ([xshift=-2.05cm,yshift=0.75cm]X_haug.south west) -- cycle;
         \draw[fill=purple!20] ([xshift=-2.1cm,yshift=-0.9cm]X_hog.north west) -- ([xshift=-0.6cm,yshift=-0.1cm]X_hog.north west) -- ([xshift=-0.6cm,yshift=-0.1cm]X_haug.south west) -- ([xshift=-2.1cm,yshift=0.8cm]X_haug.south west) -- cycle;
	\node (DEC) at ([xshift=-1.3cm,yshift=-0.1cm]X_hog.south west) {$d_{\theta}(x|z)$};

	\draw[arrow] (X_og.east) -- ([xshift=0.5cm]X_og.east);
        \draw[arrow] (X_aug.east) -- ([xshift=0.5cm]X_aug.east);
	\draw[arrow] ([xshift=0.1cm]ENC.east) -- (Z.west);
        \draw[arrow] (Z.north) -- ([xshift=-0.15cm]Laug.west);
        \draw[arrow] ([xshift=0.15cm]Laug.east) -- (Z_h.west);
	\draw[arrow] (Z.east) -- ([xshift=-0.2cm]DEC.west);
        \draw[arrow] (Z_h.east) -- ([xshift=-0.2cm]DEC.west);
	\draw[arrow] ([xshift=-0.5cm]X_hog.west) -- (X_hog.west);
        \draw[arrow] ([xshift=-0.5cm]X_hog.west) -- (X_hog.west);
        \draw[arrow] ([xshift=-0.5cm]X_haug.west) -- (X_haug.west);

        \node[] at (-2,0) {};

        \matrix [draw,below right] at ([xshift=0.5cm,yshift=0cm]current bounding box.north east) {
          \node[fill=blue!20, shape=circle,label=right:\tiny Inputs/Outputs] {}; \\
          \node[fill=purple!20, shape=circle,label=right:\tiny Trainable Networks] {}; \\
          \node[fill=red!50, shape=circle,label=right:\tiny Latent Vectors] {}; \\
        };
     
\end{tikzpicture}}
    \caption{Latent Augmentation VAE Architecture}
    \label{fig: LAVAE}
\end{figure*}
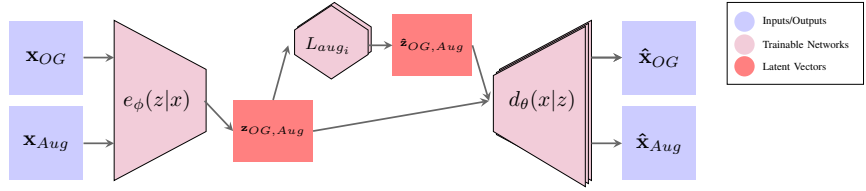

In the Latent Augmentation VAE, as seen in Figure \ref{fig: LAVAE}, we use trainable linear transforms in the latent space to learn the mappings between original and augmented latent representations resulting in a linear proxy model of the transformations applied in the image space that can be used on test data or to generate new original and augmented images. We also utilize multiple decoder heads such that one can transfer a learned latent space to a new set of augmentations by training an alternative decoder head, which can preserve certain latent space geometries and latent transform properties, improving latent augmentation performance. 

\subsection{Training LAVAE}

For our initial experiments, we focus on pairs of augmentations and their compositions (as in Figure \ref{fig: DAG}). Note there is no theoretic reason why more augmentations can not be used, but we use two for illustrating geometric properties. We train the LAVAE in three stages:

\begin{enumerate}
    \item Train the encoder/decoder and populate the latent space with the original, two types of augmentations, and their composition
    \item Learn explicit linear transformations $L_{aug_i}$ between original and augmented latent spaces
    \item Transfer trained latent space and transformations by training new decoder on any other set of augmentations
\end{enumerate}

The respective losses for each of the three stages are as follows:
\begin{gather}
    \sum_{x \in D_{aug}} \ell_{BCE}(x, \hat{x}) \quad \text{s.t.} \quad \hat{x} = \textcolor{blue}{\theta}(\textcolor{blue}{\phi}(x)) \\
    \sum_{x_0 \in D_{orig}} \sum_{k \in \mathcal{A}} (\phi(f_{aug_k}(x_0))  - \phi(x_0) \cdot \textcolor{blue}{L_{aug_k}})^2 \\
    \sum_{x_0 \in D_{orig}} \sum_{\substack{g(k) \in \mathcal{A}_T \\ k \in \mathcal{A}}} \ell_{BCE} (f_{aug_{g(k)}}(x_0), \textcolor{blue}{\theta_T}(\phi(x_0) \cdot L_{aug_k})) 
\end{gather}
where the trained parameters at each step are highlighted in blue. $\mathcal{A}, \mathcal{A_T}$ are two equal-length sets of augmentations predefined by $\{f_{aug}\}$ with $g : A \rightarrow A_T$ forming a bijective pairing between the two sets, thereby allowing the latent structure to be preserved by the transformations defined by $\mathcal{A}$. $\theta_T$ is an alternative decoder head for each new set of augmentations $\mathcal{A}_T$. In our 2-augmentation case, WLOG we can assign $\mathcal{A} = \{1, 2\}$, $\mathcal{A}_T = \{3,4\}$ where $g(1) = 3$ and $g(2) = 4$. We can also extend this formulation to any other sets of augmentations given a new mapping. Note that stage 1 also includes the KL Divergence loss to constrain the latent distribution, as described in appendix \ref{app:VAE}, with respective weights on KL and reconstruction $\lambda_{KL} = 5$ ,  $\lambda_{recon} = 1$. 

We performed experiments with non-linear latent augmentation networks, which showed slightly better performance, but lacked composability and simple invertability. We also experimented with combining training stages 1 and 2, but this degraded final performance and reconstruction. 

\subsection{Utilizing \alg{}}

Once an \alg{} is trained, there is a wide variety of uses which extend the capabilities over previous VAE methods.
There is basic reconstructions of the original, augmented, or  composed images:
\begin{gather*}
    \hat{x} = \theta(\phi(x)) \ \forall x \in D_{aug} 
\end{gather*}

We can augment the original images in the latent space:
\begin{align*}
    \hat{z_{i}} &= \phi(x_0) \cdot L_{aug_{i}}\\
    \hat{x_{i}} &= \theta(\hat{z_{i}}) \quad | \quad i\in \mathcal{A}
\end{align*}
where $x_i$ refers to $f_{aug_i}(x_0)$. 

We can go from the original images to the composed by multiplying the latent original vector by both the latent augmentation transforms, despite an explicit composition in the latent space never being trained. 
\begin{gather*}
    \hat{\mathring{z}} =  \phi(x_0) \cdot L_{aug_{1}} \cdot L_{aug_{2}}\\
    \hat{\mathring{x}} = \theta( \hat{\mathring{z}})
\end{gather*}

The reverse composition is also effective with some increased reconstruction error, indicating the latent augmentations are somewhat composable.
\begin{gather*}
    \hat{\mathring{z}}_r =  \phi(x_0) \cdot L_{aug_{2}} \cdot L_{aug_{1}}\\
    \hat{\mathring{x}} \approx \theta( \hat{\mathring{z}}_r) \\
    \hat{\mathring{z}}_r = zL_{aug_{2}}L_{aug_{1}} \approx zL_{aug_{1}}L_{aug_{2}} = \hat{\mathring{z}}\\
\end{gather*}
where $z = \phi(x_0)$. Note that this latent space property holds true for our tested augmentations even if the compositions in the image space are not equivalent (such as a $90^\circ$ rotate and flip will be different depending on the order applied). In this case, we only train the encoder and decoder, in Stage 1, on one of the compositions ($f_{aug_1}$ then $f_{aug_2}$), so the reverse composition in the latent space is not equivalent to the image space reverse composition with this process.

We can also invert the latent space transforms and go from an augmented input image to a original image, giving us `any-to-any' functionality:
\begin{align*}
    \hat{x}_{0} &= \theta(\phi(x_{i}) \cdot L_{aug_{i}}^{-1}) \quad | \quad i\in \{1,2\} \\
    \hat{x}_{0} &= \theta(\phi(\mathring{x}) \cdot (L_{aug_{1}} \cdot L_{aug_{2}})^{-1})
\end{align*}

Finally, we can run the model recursively, taking our output and running back through the network for the same or different augmentations. We find that even for general augmentations that there is some level of stability in taking the same latent augmentation over and over. 
\begin{align*}
    \hat{x}_0^{(k)} &= \theta(\phi(\hat{x}_i^{(k)}) \cdot L_{aug_{i}}) \\
    \hat{x}_i^{(k+1)} &= \theta(\phi(\hat{x}_0^{(k)}) \cdot L_{aug_{i}}) \\
    k &\in [1, n]
\end{align*}
We show results on the stability of this use-case as a recursive generator in the next section.

\section{Experiments}

All of the displayed results use the test MNIST dataset with a model trained on the training dataset. 

\subsection{LAVAE Reconstruction Results}

\begin{figure}[ht]
    \centering
    \includegraphics[width=0.4\textwidth]{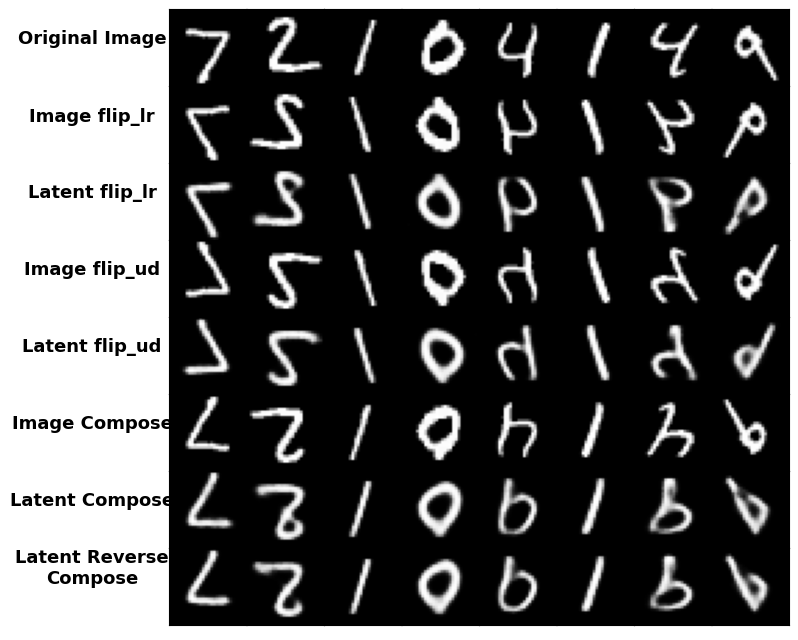}
    \caption{Eight samples of ``Flips'' latent augmentations with baseline image space augmentations for comparison}
    \label{fig:flips_recons}
\end{figure}

Figure \ref{fig:flips_recons} shows the basic and augmented reconstruction results for the ``Flips'' (flip left/right, flip up/down) augmentation pair.

\begin{figure}[ht]
    \centering
    \includegraphics[width=0.4\textwidth]{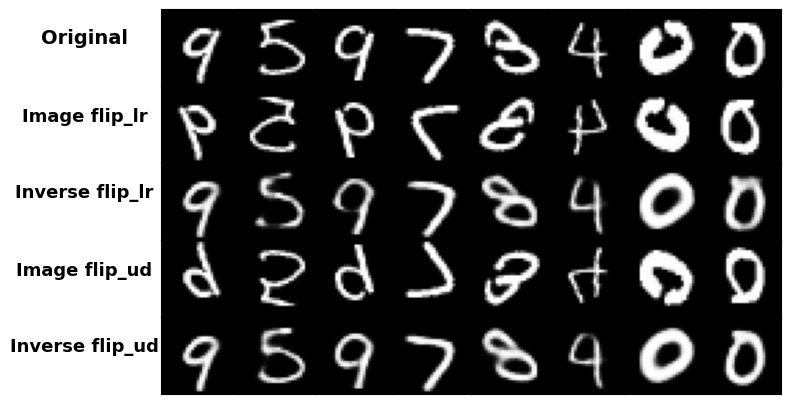}
    \caption{Eight samples of ``Flips'' latent inverse augmentations with original and augmented images (inputs) for comparison}
    \label{fig:flips_inv_recon}
\end{figure}

Figure \ref{fig:flips_inv_recon} shows the inverse reconstruction results for the Flips augmentation pair. 

\begin{figure}[ht]
    \centering
    \begin{subfigure}{0.3\textwidth}
        \centering
        \includegraphics[width=\linewidth]{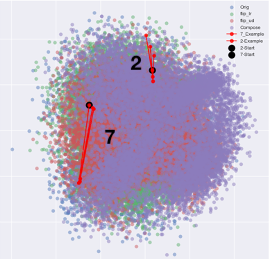}
    \end{subfigure}
    \begin{subfigure}{0.15\textwidth}
        \centering
        \includegraphics[width=.7\linewidth]{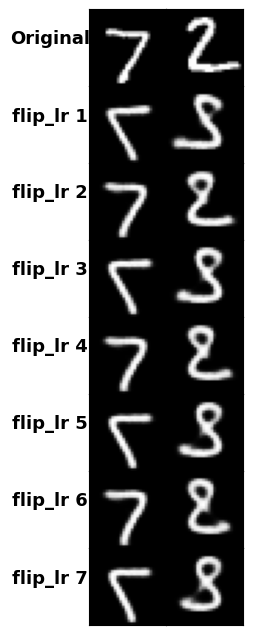}
    \end{subfigure}
    \caption{Two samples latent trajectory (2-D projection) and reconstructions of recursive flip left/right augmentation. The `7' is stable, but the `2' diverges both in latent space trajectory and reconstructions.}
    \label{fig:recursive}
\end{figure}

Figure \ref{fig:recursive} displays two examples of recursive augmentation using flip left/right, where one sample gradually deviates (from a 2 to possibly an 8), while the 7 remains relatively stable. Additionally, we illustrate a lower dimensional projection in a 2D space (using Independent Component Analysis) of the latent vectors and their corresponding ``paths'' as we repeatedly apply augmentations using \alg{}. This suggests a radius of stability around certain samples with the repeated use of augmentations.

It is worth mentioning that the \alg{} can also be applied to sample the latent space and interpolate between points. To achieve sampling, as the latent space is now divided based on the augmentation, we constructed a simple bounding box using training samples and sampled within that subspace to obtain an original image. Interpolation is simpler, as we only need to provide two test images and sample at regular intervals across all latent dimensions between the two points (16-D in this instance). Additional examples can be found in Appendix \ref{app:sample_inter}.

\subsection{LAVAE Transfer Decoders}

\begin{figure}[ht]
    \centering
    \includegraphics[width=0.4\textwidth]{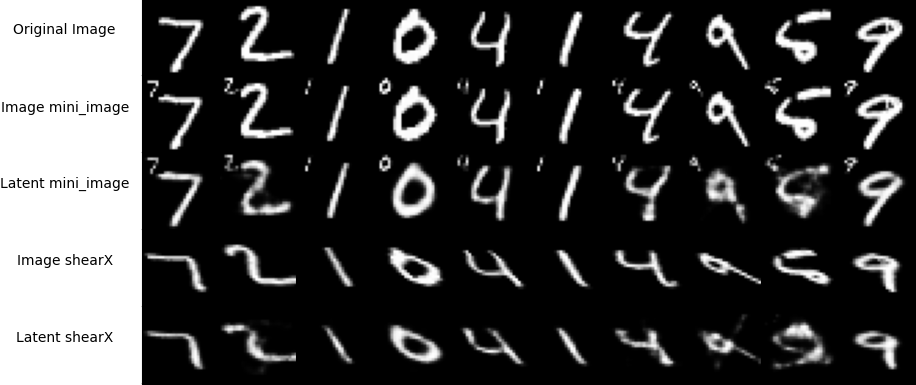}
    \caption{Eight samples of ``Flips'' latent augmentations with baseline image space augmentations for comparison}
    \label{fig:transfer_recons}
\end{figure}

\alg{} includes multiple decoder heads to enable the transfer of a trained latent space to any pair of augmentations. Figure \ref{fig:transfer_recons} shows the transfer reconstruction results from ``Flips'' to ``Nested Mini-Image, shear X-direction''. 

\begin{figure}[ht]
    \centering
    \includegraphics[width=0.4\textwidth]{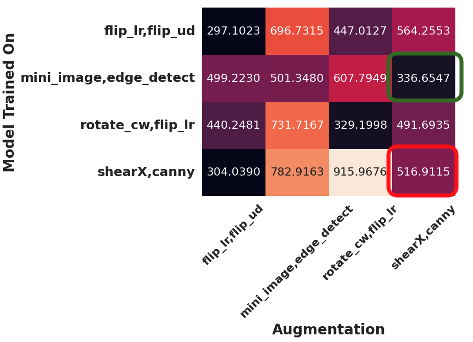}
    \caption{Initial augmentation pair choice vs. transferred augmentation MSE reconstruction error (across all augmentations)}
    \label{fig:heatmap_recons}
\end{figure}

This functionality was included because we saw that transferring to a pair of augmentations could increase the reconstruction performance over training on the augmentation pair originally. This surprising result leads us to believe certain latent space geometries, based on the choice of initial augmentations, better allow for latent augmentation reconstruction and properties such as improved composability. Figure \ref{fig:heatmap_recons} shows a heat-map matrix of reconstruction error (Mean-Squared Error in image space) with initial augmentation pair choice vs. transferred augmentation pair in which transferring from a ``Nested Mini-Image, shear X-direction'' to ``Nested Mini-Image, shear X-direction'' performs better than training on ``Nested Mini-Image, shear X-direction''. Examples of all the augmentation and more results are in appendix \ref{app:training}. 

\subsection{Conditional VAE Comparisons}
As we discussed, we realized that the Conditional VAE (CVAE) also uses a na\"ive form of latent space partitioning so we wanted to see to what extent it can do the same tasks as the \alg{}. In this case, instead of having the conditional represent the classifications, we wanted to partition just like with the \alg{} with respect to augmentation. 

\begin{figure}[ht]
    \centering
    \includegraphics[width=0.4\textwidth]{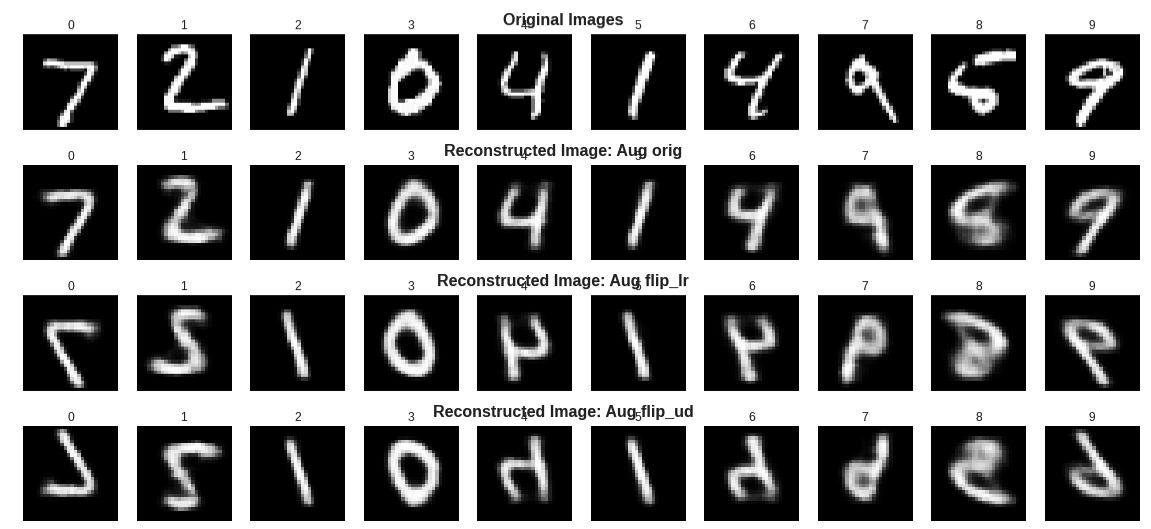}
    \caption{Initial augmentation pair choice vs. transferred augmentation MSE reconstruction error (across all augmentations)}
    \label{fig:cvae_augs}
\end{figure}

Thus, we first trained the CVAE in the traditional way where the conditional is the augmentation type. Example results are shown in \ref{fig:cvae_augs}. The decoder can reconstruct from the latent with high fidelity, but changing the conditional does not augment the image as expected. Instead, it produces a plausible image of that augmentation, but it does not preserve the uniqueness of the original image. Therefore, we can say that the conditional variables and the latent variables are ``entangled." This suggests that the CVAE cannot naively handle causally-linked images across conditionals. 

\begin{figure}[ht]
    \centering
    \includegraphics[width=0.4\textwidth]{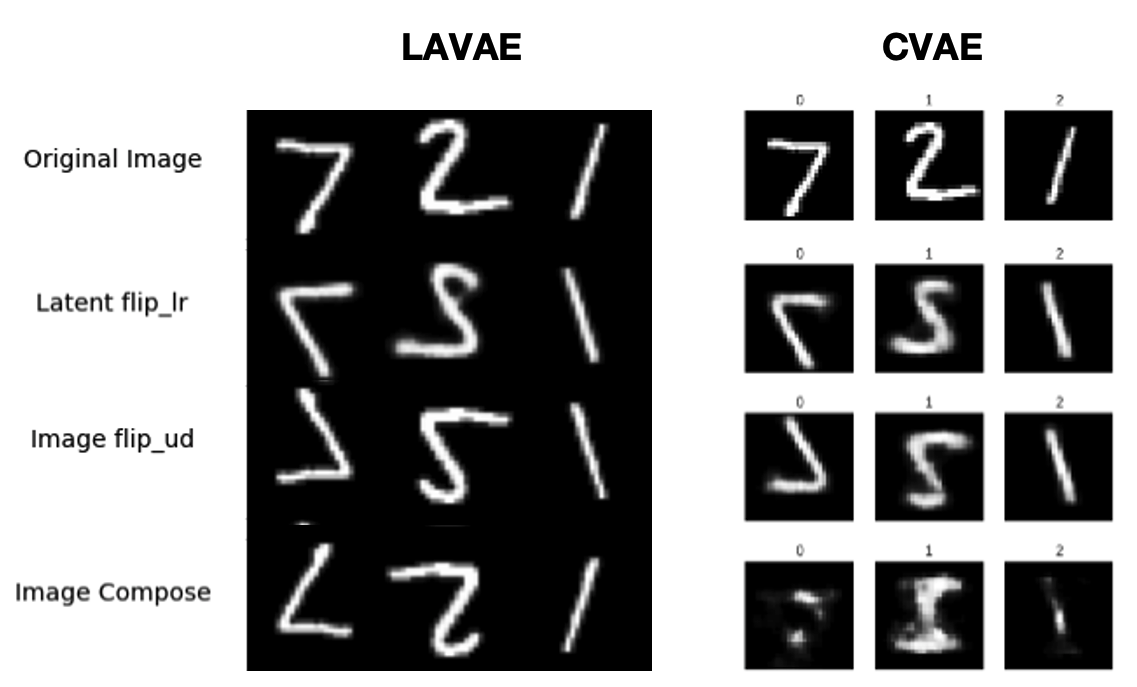}
    \caption{\alg{} vs CVAE reconstructions}
    \label{fig:cvae_vs_lavae}
\end{figure}

Fundamentally, a causal view of the CVAE would represent the following idea: any $x_i = f_{aug_i}(x_0)$ causally generated from some $x_0$ should map to the same $z$. Thus, $z$ should contain the augmentation-invariant information. Then, based on the conditional, the decoder should produce an augmented version of that image. Thus the conditional contains all the augmentation information, and we say that the augmentation and the image are disentangled.  Our second experiment attempts to show that the CVAE encoder and decoder are capable of applying augmentations to an augmentation-invariant latent space, given a similar training method to the LAVAE. However, in doing so, we can see that there is a significant hit in reconstruction loss compared to the \alg{} as can be seen in Figure \ref{fig:cvae_vs_lavae}. Furthermore, the composition property does not emerge in the same way that it does for the \alg{}. 

In Table \ref{table:LAVAE_vs_CVAE}, we compare the reconstruction errors of the \alg{} against both methods of training the CVAE on the ``Flips'' augmentation pair, showing the superior performance of \alg{}. 

\begin{table}[!t]
\renewcommand{\arraystretch}{1.1}
\newcommand*\rot{\rotatebox{0}}
\caption{``Flips'' Reconstruction Errors (MSE)}
\label{table_example}
\centering
\begin{tabular}{c|cccc|c}
\hline {Model} & \rot{Orig} & \rot{$Aug_1$} & \rot{$Aug_2$} & \rot{$\circ$} & \rot{\textbf{Total}} \\
\hline
    \scriptsize \textbf{\alg{}}                  &     \textbf{68.34} &       \textbf{75.89} &     \textbf{71.30} &          \textbf{81.57} & \textbf{297.1} \\
    \scriptsize $\textsc{CVAE}_{Trad}$    &     98.11 &        260.18 &      351.79 &          279.82 & 989.84 \\
    \scriptsize $\textsc{CVAE}_{AugInv}$  &    99.15 &        99.16 &      99.26 &           299.12   & 695.68\\
\hline
\end{tabular}
\label{table:LAVAE_vs_CVAE}\vspace{-3mm}
\end{table}

\subsection{LAVAE Latent Geometries}
\begin{figure}[ht]
    \centering
    \begin{subfigure}{0.24\textwidth}
        \centering
        \includegraphics[width=\linewidth]{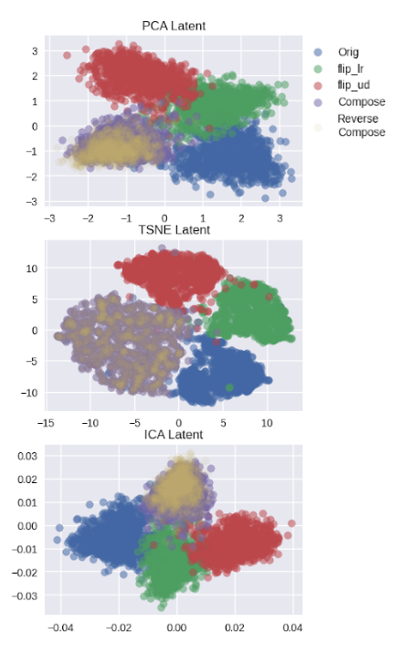}
    \end{subfigure}
    \begin{subfigure}{0.22\textwidth}
        \centering
        \includegraphics[width=\linewidth]{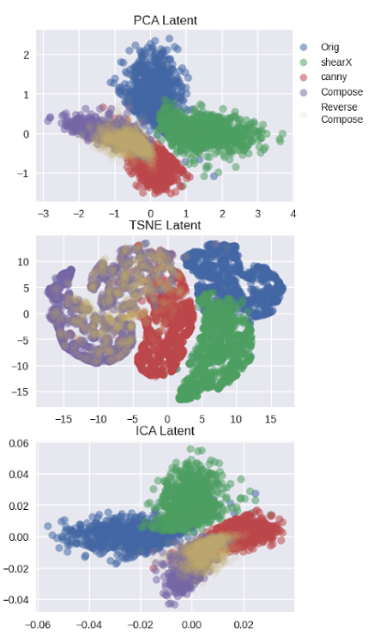}
    \end{subfigure}
    \caption{``Flips'' (left) and ``shear X-direction, canny edge-detect'' (right) 2-d projections using PCA, ICA, and T-SNE algorithms.}
    \label{fig:2d_proj}
\end{figure}

In this final section, we present a comparison of 2D projection visualizations using PCA, ICA, and T-SNE algorithms of the latent space geometries for two different pairs of augmentations, ``Flips" and ``shear X-direction,'' ``canny edge-detect," as shown in Figure \ref{fig:2d_proj}. We leave a more in-depth analysis and interpretation of the latent space geometries for future research. However, we would like to point out that while symmetries seem to exist across augmentation pairs, the separation between regions and the relative areas of regions vary significantly across pairs. Additional images can be found in Appendix \ref{app:latent_geo}.
\section{Conclusion}

Data augmentation is a critical technique in deep-learning image models that can enhance generalization. In this paper, we have introduced a novel approach for \textit{latent} image augmentation using a Variational Autoencoder (VAE) architecture. This method facilitates the combination of multiple augmentation techniques and offers greater control and interpretability of the latent space. Our experiments on the MNIST dataset have shown that our proposed approach outperforms comparable models in both flexibilities of usage and performance. Furthermore, our method provides new insights into the underlying structure of the data and the relationship between different augmentations.

By treating augmentations as image-space priors instead of simply randomizing data, we can constrain the VAE's information bottleneck and learn a low-dimensional proxy for the augmentation model. For future work, we plan to explore the combination of the CVAE and the LAVAE, such as producing latent augmentations per class conditional, changing the one-hot conditional to continuous augmentations, and having augmentations operate only on the conditional embedding space. Additionally, we aim to apply our approach to other datasets, including those where the image model might not be known, such as 2D to 3D reconstruction or perspective shift tasks.
\bibliographystyle{IEEEtran}
\bibliography{IEEEabrv,lavae}

\begin{thebibliography}{10}
\providecommand{\url}[1]{#1}
\csname url@samestyle\endcsname
\providecommand{\newblock}{\relax}
\providecommand{\bibinfo}[2]{#2}
\providecommand{\BIBentrySTDinterwordspacing}{\spaceskip=0pt\relax}
\providecommand{\BIBentryALTinterwordstretchfactor}{4}
\providecommand{\BIBentryALTinterwordspacing}{\spaceskip=\fontdimen2\font plus
\BIBentryALTinterwordstretchfactor\fontdimen3\font minus
  \fontdimen4\font\relax}
\providecommand{\BIBforeignlanguage}[2]{{%
\expandafter\ifx\csname l@#1\endcsname\relax
\typeout{** WARNING: IEEEtran.bst: No hyphenation pattern has been}%
\typeout{** loaded for the language `#1'. Using the pattern for}%
\typeout{** the default language instead.}%
\else
\language=\csname l@#1\endcsname
\fi
#2}}
\providecommand{\BIBdecl}{\relax}
\BIBdecl

\bibitem{sohn_cvae:2015}
\BIBentryALTinterwordspacing
K.~Sohn, H.~Lee, and X.~Yan, ``Learning structured output representation using
  deep conditional generative models,'' in \emph{Advances in Neural Information
  Processing Systems}, C.~Cortes, N.~Lawrence, D.~Lee, M.~Sugiyama, and
  R.~Garnett, Eds., vol.~28.\hskip 1em plus 0.5em minus 0.4em\relax Curran
  Associates, Inc., 2015. [Online]. Available:
  \url{https://proceedings.neurips.cc/paper/2015/file/8d55a249e6baa5c06772297520da2051-Paper.pdf}
\BIBentrySTDinterwordspacing

\bibitem{aaron_vqvae:2017}
\BIBentryALTinterwordspacing
A.~van~den Oord, O.~Vinyals, and K.~Kavukcuoglu, ``Neural discrete
  representation learning,'' \emph{CoRR}, vol. abs/1711.00937, 2017. [Online].
  Available: \url{http://arxiv.org/abs/1711.00937}
\BIBentrySTDinterwordspacing

\bibitem{Su_fvae:2018}
\BIBentryALTinterwordspacing
J.~Su and G.~Wu, ``f-vaes: Improve vaes with conditional flows,'' \emph{CoRR},
  vol. abs/1809.05861, 2018. [Online]. Available:
  \url{http://arxiv.org/abs/1809.05861}
\BIBentrySTDinterwordspacing

\bibitem{rombach_diff:2022}
R.~Rombach, A.~Blattmann, D.~Lorenz, P.~Esser, and B.~Ommer, ``High-resolution
  image synthesis with latent diffusion models,'' in \emph{Proceedings of the
  IEEE/CVF Conference on Computer Vision and Pattern Recognition}, 2022, pp.
  10\,684--10\,695.

\bibitem{yang_causalvae:2021}
\BIBentryALTinterwordspacing
M.~Yang, F.~Liu, Z.~Chen, X.~Shen, J.~Hao, and J.~Wang,
  ``\BIBforeignlanguage{en}{{CausalVAE}: {Disentangled} {Representation}
  {Learning} via {Neural} {Structural} {Causal} {Models}},''
  \emph{\BIBforeignlanguage{en}{arXiv:2004.08697 [cs, stat]}}, Mar. 2021,
  arXiv: 2004.08697. [Online]. Available: \url{http://arxiv.org/abs/2004.08697}
\BIBentrySTDinterwordspacing

\bibitem{kocaoglu_causalgan:2017}
\BIBentryALTinterwordspacing
M.~Kocaoglu, C.~Snyder, A.~G. Dimakis, and S.~Vishwanath,
  ``\BIBforeignlanguage{en}{{CausalGAN}: {Learning} {Causal} {Implicit}
  {Generative} {Models} with {Adversarial} {Training}},''
  \emph{\BIBforeignlanguage{en}{arXiv:1709.02023 [cs, math, stat]}}, Sep. 2017,
  arXiv: 1709.02023. [Online]. Available: \url{http://arxiv.org/abs/1709.02023}
\BIBentrySTDinterwordspacing

\bibitem{bhat_debias:2022}
\BIBentryALTinterwordspacing
S.~Bhat, J.~Jiang, O.~Pooladzandi, and G.~Pottie, ``De-biasing generative
  models using counterfactual methods,'' 2022. [Online]. Available:
  \url{https://arxiv.org/abs/2207.01575}
\BIBentrySTDinterwordspacing

\bibitem{bhat_hypothesis:2022}
\BIBentryALTinterwordspacing
S.~G. Bhat, O.~Pooladzandi, J.~Jiang, and G.~Pottie, ``Hypothesis testing using
  causal and causal variational generative models,'' in \emph{NeurIPS 2022
  Workshop on Synthetic Data for Empowering ML Research}, 2022. [Online].
  Available: \url{https://openreview.net/forum?id=diB-RsDQJeG}
\BIBentrySTDinterwordspacing

\bibitem{breugel_decaf:2021}
\BIBentryALTinterwordspacing
B.~van Breugel, T.~Kyono, J.~Berrevoets, and M.~van~der Schaar, ``{DECAF}:
  Generating fair synthetic data using causally-aware generative networks,'' in
  \emph{Advances in Neural Information Processing Systems}, A.~Beygelzimer,
  Y.~Dauphin, P.~Liang, and J.~W. Vaughan, Eds., 2021. [Online]. Available:
  \url{https://openreview.net/forum?id=XN1M27T6uux}
\BIBentrySTDinterwordspacing

\bibitem{Sinha_negAug:2021}
\BIBentryALTinterwordspacing
A.~Sinha, K.~Ayush, J.~Song, B.~Uzkent, H.~Jin, and S.~Ermon, ``Negative data
  augmentation,'' 2021. [Online]. Available:
  \url{https://arxiv.org/abs/2102.05113}
\BIBentrySTDinterwordspacing

\end{thebibliography}

\newpage
\onecolumn
\appendix

\subsection{Variational Autoencoders (VAEs)} \label{app:VAE}

The VAE framework is based on the principle of maximum likelihood estimation , where the goal is to maximize the likelihood of the training data under the model. However, in order to make the optimization tractable, the VAE introduces a variational lower bound on the log likelihood, which can be written as:

\begin{align}
\mathcal{L}(\theta,\phi;x^{(i)}) &= \mathbb{E}_{e_{\phi}(z|x^{(i)})}\left[\log d_\theta(x^{(i)}|z)\right]  -\text{KL}\left(e_\phi(z|x^{(i)})||m_\theta(z)\right)  \\
&= \mathbb{E}_{e_{\phi}(z|x^{(i)})}\left[\log d_\theta(x^{(i)}|z)\right]
- \int e_\phi(z|x^{(i)}) \log \frac{e_\phi(z|x^{(i)})}{m_\theta(z)} dz
\end{align}

where $x^{(i)}$ is a single training example, and $\theta$ and $\phi$ are the parameters of the decoder and encoder, respectively. The first term in the lower bound, $\mathbb{E}_{e_{\phi}(z|x^{(i)})}\left[\log d_\theta(x^{(i)}|z)\right]$, is known as the reconstruction loss, and it measures the difference between the reconstructed data and the original data. The second term, $\text{KL}\left(e_\phi(z|x^{(i)})||m_\theta(z)\right)$, is known as the KL divergence, and it measures the difference between the approximate posterior distribution and the latent distribution. The first term is the decoding error (the classic rate-distortion theory), and the second term is the extra rate for coding $z$ assuming marginal pdf $m_\theta(z)$.

\subsection{Conditional VAE (CVAE)} \label{app: CVAE}

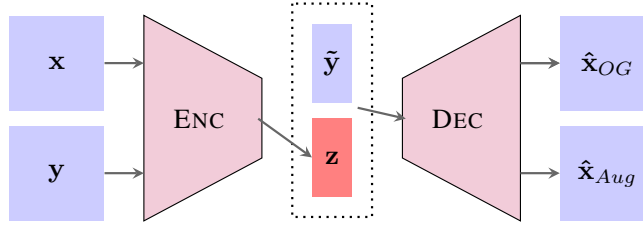
\begin{figure}[h]
\centering
    \resizebox{0.48\textwidth}{!}{

\begin{tikzpicture}

	\node[fill=blue!20, minimum width=1.2cm, minimum height=1.2cm] (X_og) at (0,1) {$\mathbf x$};
        \node[fill=blue!20, minimum width=1.2cm, minimum height=1.2cm] (y) at (0,-0.4) {$\mathbf y$};

	\draw[fill=purple!20] ([xshift=0.5cm]X_og.north east) -- ([xshift=2cm,yshift=-0.8cm]X_og.north east) -- ([xshift=2cm,yshift=0.8cm]y.south east) -- ([xshift=0.5cm]y.south east) -- cycle; 
	\node (ENC) at ([xshift=1.2cm,yshift=-0.1cm]X_og.south east) {\textsc{Enc}};

	\node[fill=red!50, minimum width=0.5cm, minimum height=1.0cm] (Z) at (3.5cm,-0.2cm) {$\mathbf z$};

        

        \node[fill=blue!20, minimum width=0.5cm, minimum height=1.0cm] (y_latent) at (3.5cm,1cm) {$\mathbf {\tilde{y}}$};

        \draw[thick,dotted] ($(y_latent.north west)+(-0.25,0.25)$)  rectangle ($(Z.south east)+(0.25,-0.25)$);
        \node (latent) at ($(Z.north) + (0,0.125)$) {};

	\node[fill=blue!20, minimum width=1.2cm, minimum height=1.2cm] (X_hog) at (7,1) {$\mathbf{\hat{x}}_{OG}$};
        \node[fill=blue!20, minimum width=1.2cm, minimum height=1.2cm] (X_haug) at (7,-0.4) {$\mathbf{\hat{x}}_{Aug}$};

	\draw[fill=purple!20] ([xshift=-2cm,yshift=-0.8cm]X_hog.north west) -- ([xshift=-0.5cm]X_hog.north west) -- ([xshift=-0.5cm]X_haug.south west) -- ([xshift=-2cm,yshift=0.8cm]X_haug.south west) -- cycle;
	\node (DEC) at ([xshift=-1.3cm,yshift=-0.1cm]X_hog.south west) {\textsc{Dec}};

	\draw[arrow] (X_og.east) -- ([xshift=0.5cm]X_og.east);
        \draw[arrow] (y.east) -- ([xshift=0.5cm]y.east);
	\draw[arrow] ([xshift=0.3cm]ENC.east) -- (Z.west);
	\draw[arrow] ([xshift=0.2cm]latent.east) -- ([xshift=-0.2cm]DEC.west);
	\draw[arrow] ([xshift=-0.5cm]X_hog.west) -- (X_hog.west);
        \draw[arrow] ([xshift=-0.5cm]X_hog.west) -- (X_hog.west);
        \draw[arrow] ([xshift=-0.5cm]X_haug.west) -- (X_haug.west);


     
\end{tikzpicture}}
    \caption{Conditional VAE Architecture}
    \label{fig: CVAE}
\end{figure}

Let $y$ be the conditional input, and let $x$ be the data that we want to generate. The encoder in the CVAE takes both $y$ and $x$ as input and maps them to the latent space $z$ as seen in Figure \ref{fig: CVAE}. Specifically, the encoder outputs a mean vector $\mu_{\phi}(z|y,x)$ and a variance vector $\sigma^2_{\phi}(z|y,x)$ that parameterize a Gaussian distribution $e_{\phi}(z|y,x)$ over the latent variables $z$. The decoder takes the latent variables $z$ and the conditional input $y$ as input and maps them to the data space. Specifically, the decoder outputs a mean vector $\mu_{\theta}(x|y,z)$ that parameterizes a Gaussian distribution $d_{\theta}(x|y,z)$ over the data variables $x$. 

The lower bound for the CVAE can be derived in a similar way to the VAE, by introducing a variational lower bound on the log likelihood. The conditional version of the lower bound is given by:

\begin{align}
\mathcal{L}(\theta,\phi;x^{(i)},y^{(i)}) &= \mathbb{E}{e{\phi}(z|y^{(i)},x^{(i)})}\left[\log d_\theta(x^{(i)}|y^{(i)},z)\right] 
- \text{KL}\left(e_\phi(z|y^{(i)},x^{(i)})||m_\theta(z|y^{(i)})\right)  \\
&= \mathbb{E}{e{\phi}(z|y^{(i)},x^{(i)})}\left[\log d_\theta(x^{(i)}|y^{(i)},z)\right] 
- \int e_\phi(z|y^{(i)},x^{(i)}) \log \frac{e_\phi(z|y^{(i)},x^{(i)})}{m_\theta(z|y^{(i)})} dz
\end{align}

where $x^{(i)}$ and $y^{(i)}$ are a single training example and its corresponding condition, respectively. The first term in the lower bound, $\mathbb{E}{e{\phi}(z|y^{(i)},x^{(i)})}\left[\log d_\theta(x^{(i)}|y^{(i)},z)\right]$, measures the reconstruction loss, i.e., the difference between the reconstructed data and the original data given the condition $y^{(i)}$. The second term, $\text{KL}\left(e_\phi(z|y^{(i)},x^{(i)})||m_\theta(z|y^{(i)})\right)$, measures the KL divergence between the approximate posterior distribution $e_\phi(z|y^{(i)},x^{(i)})$ and the prior distribution $m_\theta(z|y^{(i)})$ over the latent variables $z$ given the condition $y^{(i)}$.

The CVAE can be used for a variety of tasks, such as image and text generation, where the conditional input $y$ corresponds to a class label and the data $x$ is an image  respectively. In the CVAE framework, the encoder network takes both the input data $x$ and the conditional input $y$ as inputs and produces the approximate posterior distribution $e_{\phi}(z|x,y)$ over the latent variable $z$. Similarly, the decoder network takes both $z$ and $y$ as inputs and produces the reconstructed output $d_{\theta}(x|z,y)$.

The objective function for the CVAE can be derived from the VAE's lower bound by conditioning on the conditional input $y$:

\begin{align}
\mathcal{L}_{\text{CVAE}}(\theta,\phi;x^{(i)},y^{(i)}) &= \mathbb{E}{e_{\phi}(z|x^{(i)},y^{(i)})}\left[\log d_\theta(x^{(i)}|z,y^{(i)})\right]  -\text{KL}\left(e_\phi(z|x^{(i)},y^{(i)})||m_\theta(z|y^{(i)})\right) \\
&= \mathbb{E}{e{\phi}(z|x^{(i)},y^{(i)})}\left[\log d_\theta(x^{(i)}|z,y^{(i)})\right] 
- \int e_\phi(z|x^{(i)},y^{(i)}) \log \frac{e_\phi(z|x^{(i)},y^{(i)})}{m_\theta(z|y^{(i)})} dz
\end{align}

where $y^{(i)}$ is the conditional input for the $i$-th training example and $\theta$ and $\phi$ are the parameters of the decoder and encoder, respectively. The first term in the lower bound, $\mathbb{E}{e{\phi}(z|x^{(i)},y^{(i)})}\left[\log d_\theta(x^{(i)}|z,y^{(i)})\right]$, measures the difference between the reconstructed output and the original output, given the input and the conditional input. The second term, $\text{KL}\left(e_\phi(z|x^{(i)},y^{(i)})||m_\theta(z|y^{(i)})\right)$, measures the difference between the approximate posterior distribution and the prior distribution of the latent variable, given the conditional input.

\subsection{Architecture and Training}\label{app:training}

\begin{itemize}
    \item Model: (Total Parameter Count: )
    Encoder/Decoder:  (Parameter Count: )
    \begin{itemize}
        \item Encoder: 2 conv layers followed by a fully-connected layer. For CVAE, append one-hot representation of the augmentation class to the features before the FC layer. 
        \item Latent Space: 16 dimensions. 
        \item Decoder: Fully-connected layer followed by 2 ConvTranspose layers. For the CVAE, again append one-hot representation of the augmentation class to the latent space inputs. For LAVAE, each pair of augmentations gets its own decoder head. 
        \item For LAVAE, $L_{aug_i}$: 2 16 $\times$ 16 linear matrices. Will be used regardless of which decoder is being used. 
    \end{itemize}
    \item Optimizer: Adabelief, one for Encoder/Decoder, one for latent augmentation networks, and one for each additional decoder head
    \begin{itemize}
        \item learning-rate: 0.0001
        \item epsilon=1e-16, betas=(0.9,0.999)
    \end{itemize}
    \item Training Epochs
    \begin{itemize}
        \item Encoder/Decoder - 100
        \item Latent Augmentation Networks - 60
        \item Additional Decoders - 100
    \end{itemize}
    \item Training Parameters
    \begin{itemize}
        \item batch-size 64
    \end{itemize}
\end{itemize}

\begin{figure}[htb!]
\centering
    \includegraphics[width=0.25\linewidth]{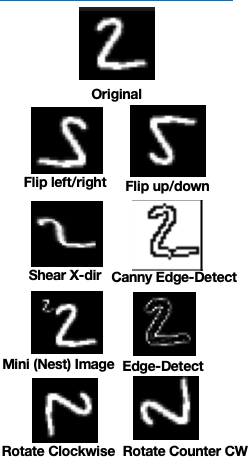}
    \caption{All Augmentations Visualized}
\end{figure}

\FloatBarrier
\subsection{Sampling and Interpolation}\label{app:sample_inter}

\begin{figure}[htb!]
\centering
    \includegraphics[width=0.25\linewidth]{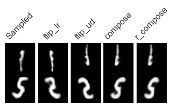}
    \caption{Sampled digits and augmentations via bounding box method}
\end{figure}

\begin{figure}[htb!]
\centering
    \includegraphics[width=0.7\linewidth]{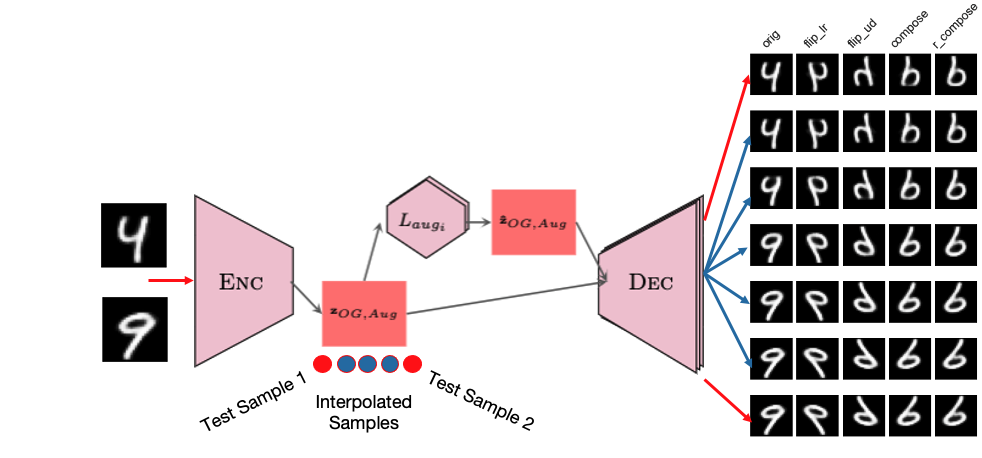}
    \caption{Interpolating between two test samples (top and bottom row) with augmentations}
\end{figure}

\FloatBarrier
\subsection{Additional Augmentation Reconstructions}

\begin{figure}[htb!]
\centering
    \includegraphics[width=0.80\linewidth]{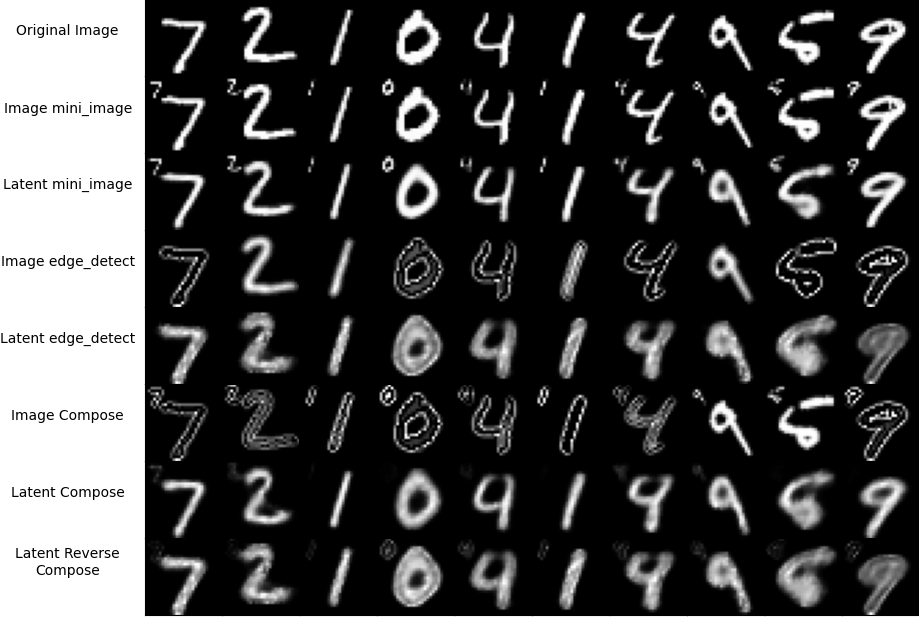}
    \caption{``Nested Mini-Image, Edge-Detect'' Reconstructions}
\end{figure}

\begin{figure}[htb!]
\centering
    \includegraphics[width=0.80\linewidth]{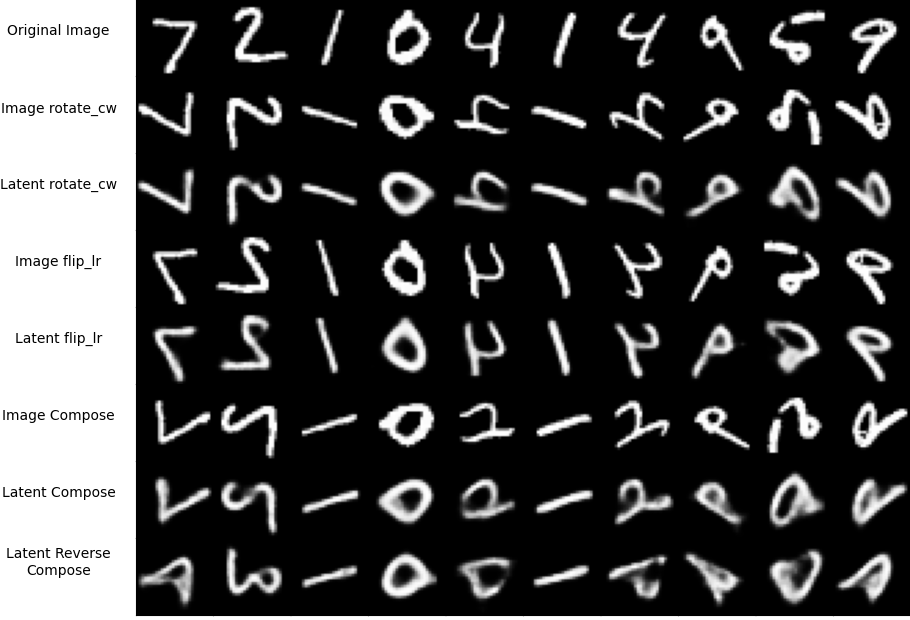}
    \caption{``$90\deg$ Clockwise Rotation, Flip left/right'' Reconstructions}
\end{figure}

\begin{figure}[htb!]
\centering
    \includegraphics[width=0.80\linewidth]{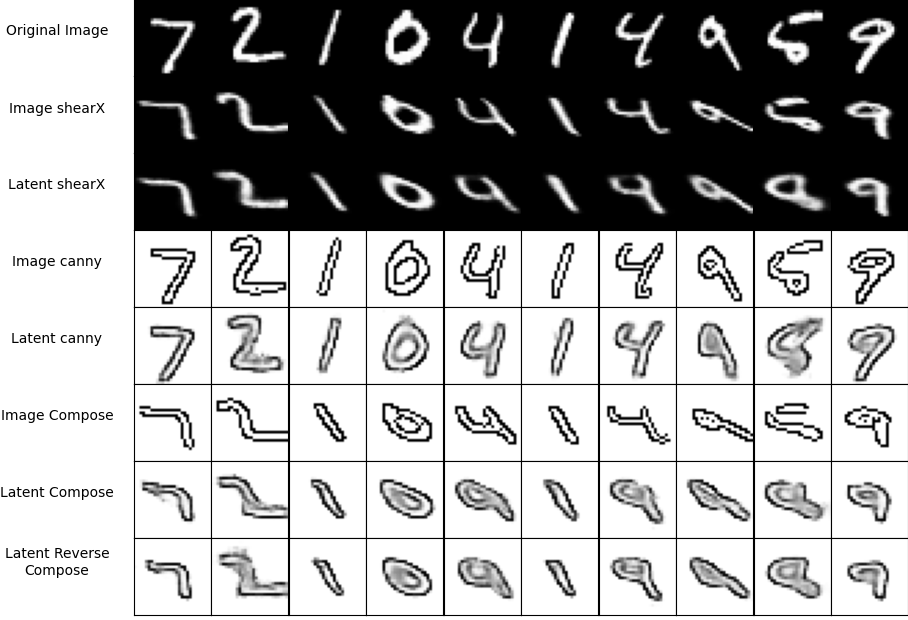}
    \caption{``X-direction shear, Canny edge-detect'' Reconstructions}
\end{figure}

\FloatBarrier
\subsection{Additional Transfer Decoder Results}

\begin{figure}[htb!]
\centering
    \includegraphics[width=0.55\linewidth]{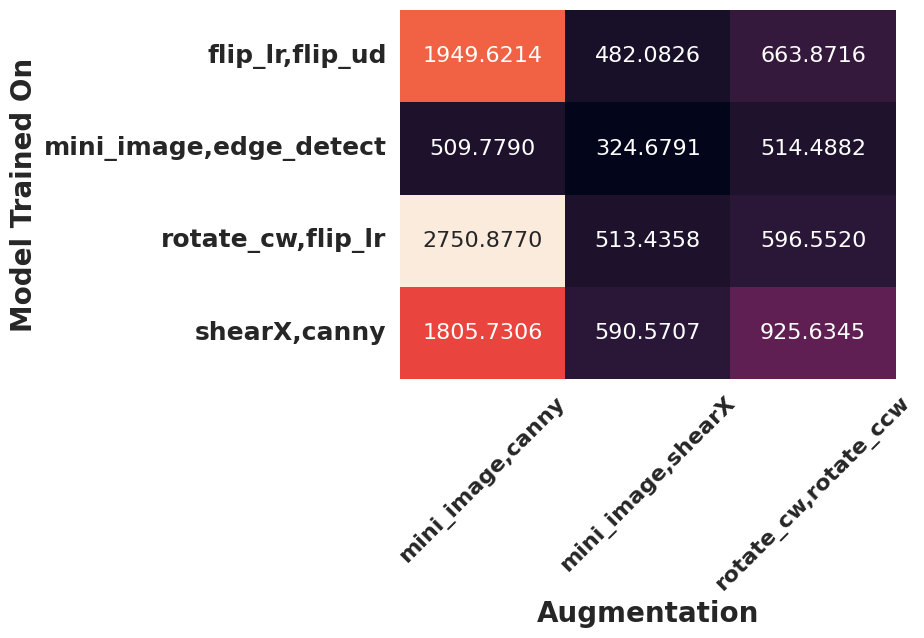}
    \caption{Additional transfer decoder head results}
\end{figure}

\FloatBarrier
\subsection{Latent Space Geometries}\label{app:latent_geo}

\begin{figure}[htb!]
\centering
    \includegraphics[width=0.80\linewidth]{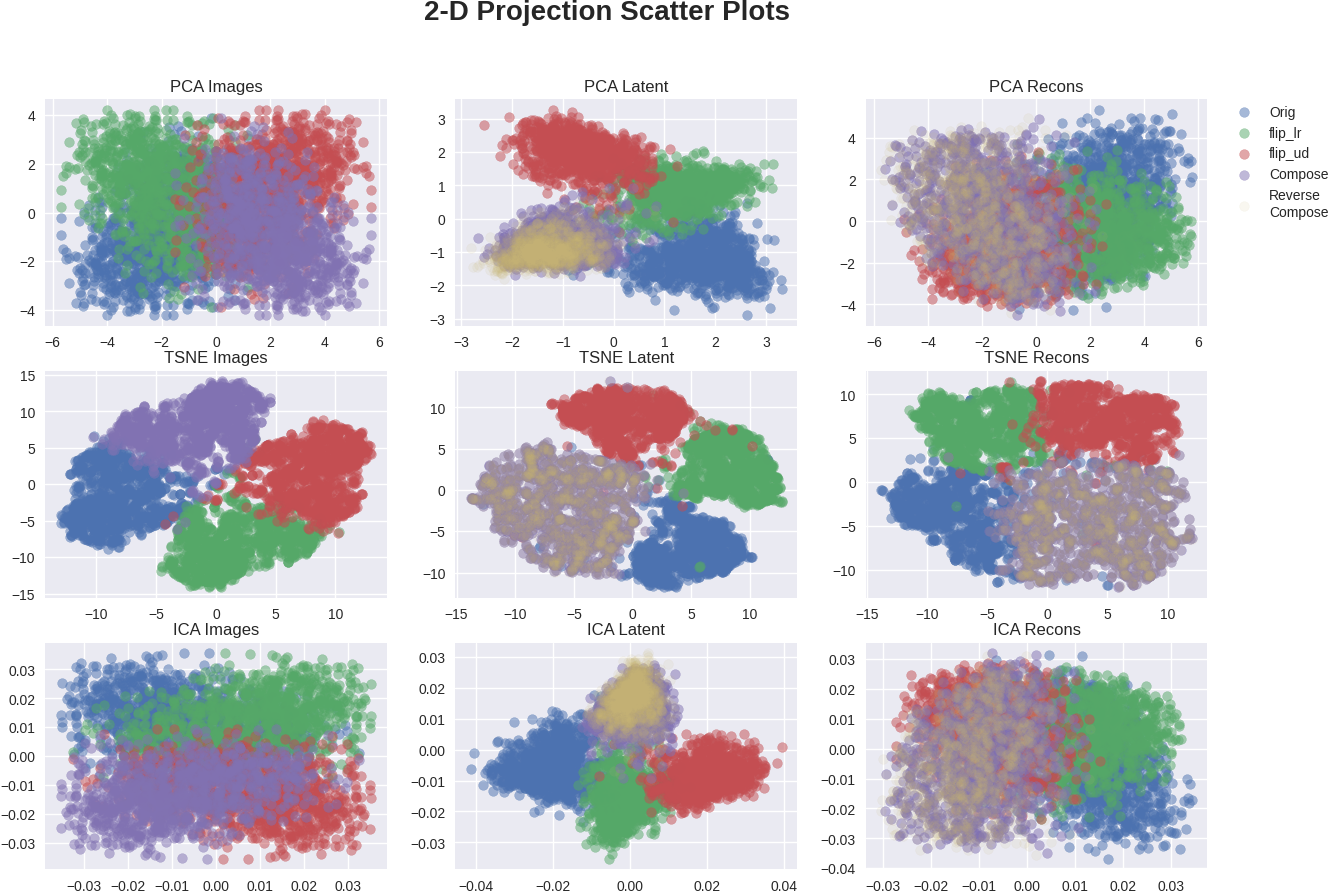}
    \caption{``Flips'' Image, Latent, and Reconstructions Image 2-D Projections}
\end{figure}

\begin{figure}[htb!]
\centering
    \includegraphics[width=0.80\linewidth]{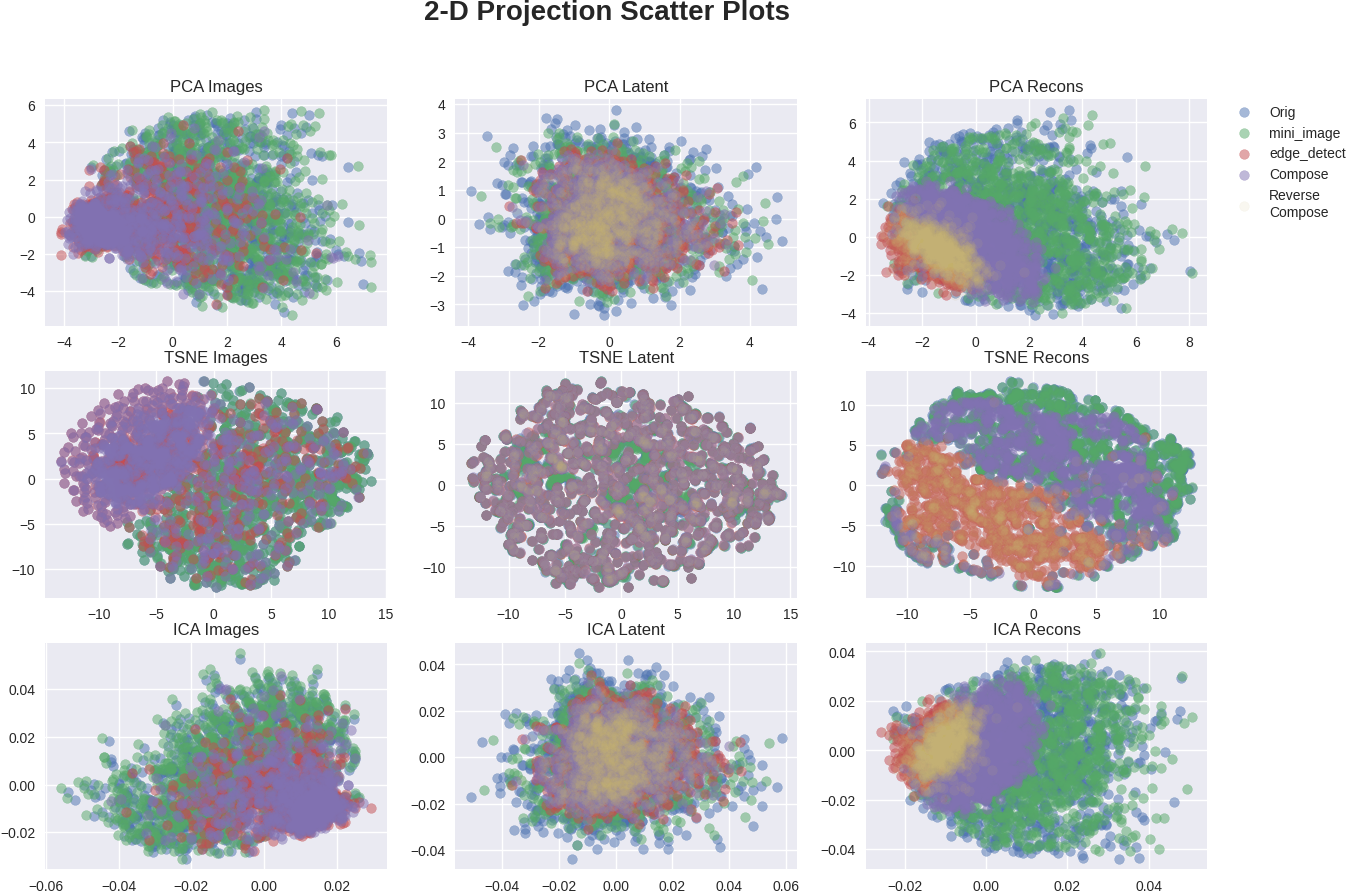}
    \caption{``Nested Mini-Image, Edge-Detect'' Image, Latent, and Reconstructions Image 2-D Projections}
\end{figure}

\begin{figure}[htb!]
\centering
    \includegraphics[width=0.80\linewidth]{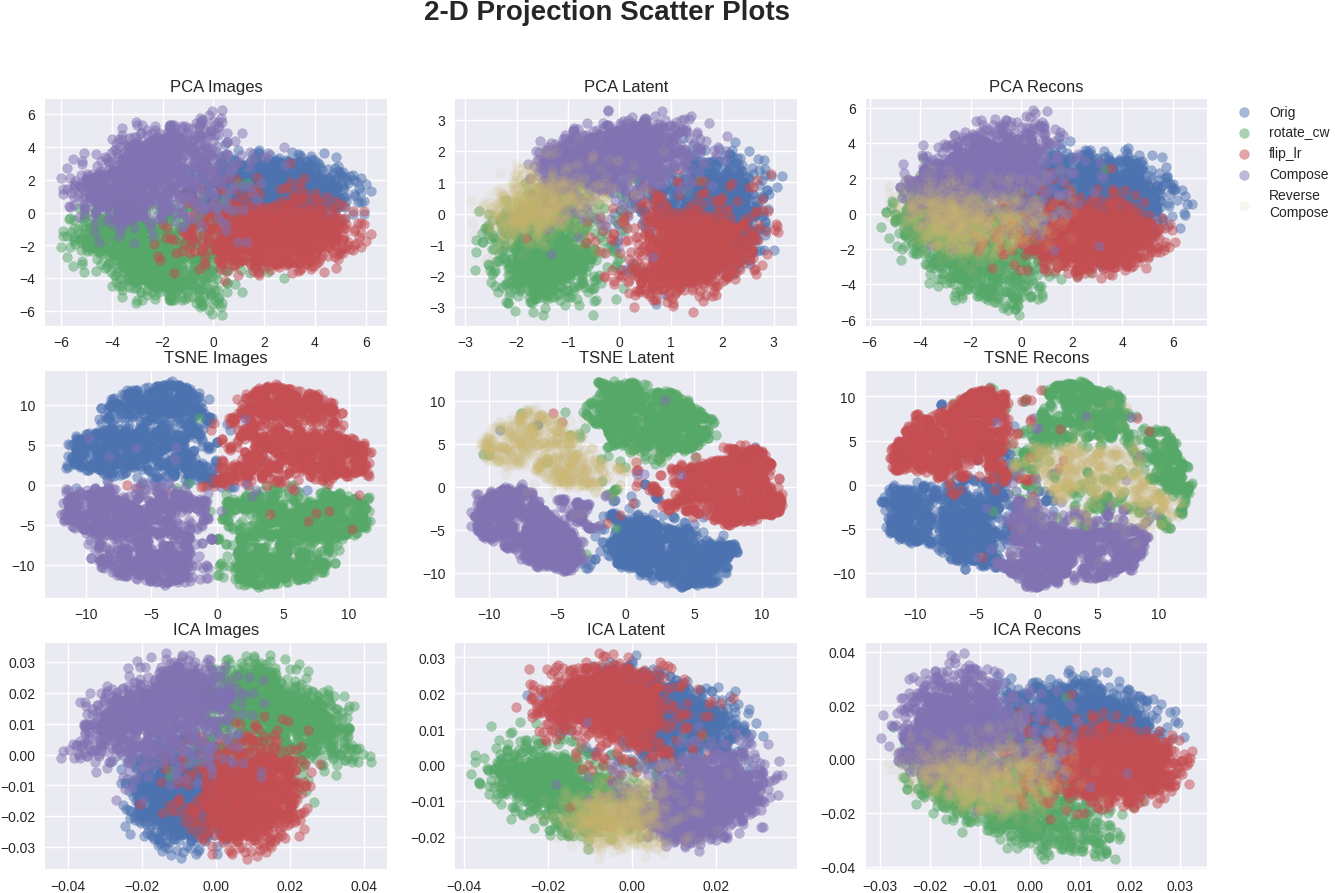}
    \caption{``$90^\circ$ Clockwise Rotation, Flip left/right'' Image, Latent, and Reconstructions Image 2-D Projections}
\end{figure}

\begin{figure}[htb!]
\centering
    \includegraphics[width=0.80\linewidth]{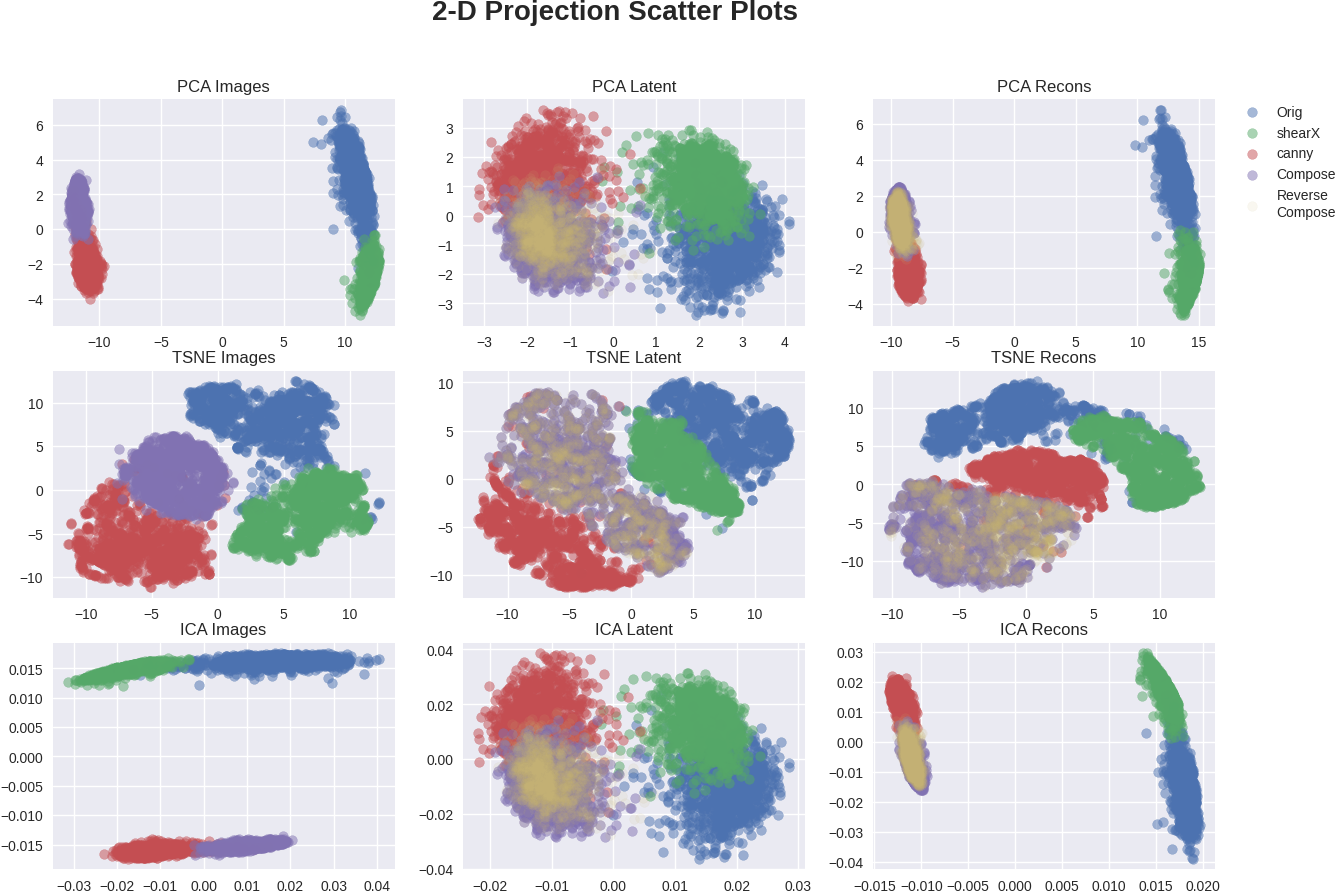}
    \caption{``X-direction shear, Canny edge-detect'' Image, Latent, and Reconstructions Image 2-D Projections}
\end{figure}

\end{document}